\documentclass[runningheads]{llncs}
\usepackage{graphicx}
\usepackage{multirow}
\usepackage{xurl}
\usepackage{listings}
\usepackage{paralist}

\usepackage{cite}
\usepackage{hyperref}
\begin{document}
%
% \title{Comparative Analysis of Artificial Intelligence for Indian Legal Question Answering (AILQA) Using Different Retrieval and QA Models}

\title{Legal Question-Answering in the Indian Context: Efficacy, Challenges, and Potential of Modern AI Models}
%
% \titlerunning{Comparative Analysis of AI for Indian Legal QA (AILQA)}
\titlerunning{Legal Question-Answering in the Indian Context}
% If the paper title is too long for the running head, you can set
% an abbreviated paper title here

\author{Shubham Kumar Nigam\inst{1} \and
Shubham Kumar Mishra\inst{1} \and
Ayush Kumar Mishra\inst{1} \and
Noel Shallum\inst{2} \and
Arnab Bhattacharya\inst{1}}
% % %
\authorrunning{Shubham Kumar Nigam et al.}
% % % First names are abbreviated in the running head.
% % % If there are more than two authors, 'et al.' is used.
% % %
\institute{Indian Institute of Technology Kanpur, India \and
Symbiosis Law School Pune, India \\
\email{\{sknigam,skmishra20,ayushkm20,arnabb\}@iitk.ac.in}\\
\email{noelshallum@gmail.com}}

\maketitle              % typeset the header of the contribution
\begin{abstract}

Legal QA platforms bear the promise to metamorphose the manner in which legal experts engage with jurisprudential documents. In this exposition, we embark on a comparative exploration of contemporary AI frameworks, gauging their adeptness in catering to the unique demands of the Indian legal milieu, with a keen emphasis on Indian Legal Question Answering (AILQA). Our discourse zeroes in on an array of retrieval and QA mechanisms, positioning the OpenAI GPT model as a reference point. The findings underscore the proficiency of prevailing AILQA paradigms in decoding natural language prompts and churning out precise responses. The ambit of this study is tethered to the Indian criminal legal landscape, distinguished by its intricate nature and associated logistical constraints. To ensure a holistic evaluation, we juxtapose empirical metrics with insights garnered from seasoned legal practitioners, thereby painting a comprehensive picture of AI's potential and challenges within the realm of Indian legal QA.

% \keywords{First keyword  \and Second keyword \and Another keyword.}

\end{abstract}
\section{Introduction}
Question Answering (QA) \cite{choi2018quac}, \cite{allam2012question}, \cite{allam2012question, choi2018quac} is an artificial intelligence (AI) task designed to provide answers to queries in a natural language, similar to how humans do. QA systems typically utilize Natural Language Processing (NLP) methods to understand the question's meaning and then employ techniques like machine learning and information retrieval to find the most suitable answers from a vast dataset. Question-answering (QA) systems have gained significant attention in recent years due to their potential to transform the way we interact with information. The ability to extract relevant information from vast amounts of unstructured data has many applications, including answering legal questions. Legal QA systems have the potential to assist legal professionals in the interpretation and application of legal principles, improve the efficiency of legal research, and ultimately enhance the delivery of justice. QA systems have already shown promising results in various domains, including healthcare, customer service, and education.

In the realm of QA, deep learning methods \cite{weston2015towards} enhance the system's ability to comprehend question meaning and identify the most appropriate answer from a large dataset. Over the past few years, deep learning has gained significant popularity and has been instrumental in building state-of-the-art QA systems that deliver highly accurate answers for a wide range of questions and the need for machines to understand human language. Prominent examples of question-answering systems employing deep learning include Generative Pretrained Transformer 3 (GPT-3) and Google's BERT\cite{devlin2018bert, qu2019bert, wang2019multi, kassner2020bert}. QA systems have already shown promising results in various domains, including healthcare, customer service, and education.

However, building effective legal QA systems poses several challenges, such as dealing with complex and diverse legal language, recognizing the context of legal cases, and understanding the nuances of legal reasoning. These challenges are particularly significant in the Indian legal domain, which has a unique legal system and language that differs significantly from other legal systems worldwide.

Our paper investigates several embedding and QA model combinations for Indian legal question answering, providing a more comprehensive evaluation of the task's challenges and potential solutions. Our work leverages LLM-based Generative Pretrained Transformer (GPT-3 Davinci model) \cite{brown2020language}, a state-of-the-art language model trained by OpenAI, which has revolutionized the field of natural language processing. Specifically, we evaluate eight different experimental settings, including four combinations of embedding and QA models, ChatGPT answers, and retrieval of relevant information from the database using a syntactic-based algorithm (BM25) followed by an OpenAI embedding and QA model.

We evaluate the performance of our models using popular syntactic evaluation metrics, such as Rouge 1, Rouge 2, Rouge L, BLEU score, and semantic evaluation metrics using MPNET score. Additionally, we conduct legal expert evaluations for each category of the predictions using a rating system ranging from 1 to 5.

We focus on criminal cases in our study due to resource and time constraints associated with hiring legal experts to evaluate other types of legal cases, such as civil law cases or family law cases. However, we believe our results provide valuable insights into the potential of embedding and QA models in the Indian legal domain and can be extended to other legal domains with appropriate evaluation mechanisms. Upon acceptance of the paper, we intend to make our code and dataset publicly available.

Our study provides a comprehensive evaluation of the challenges and potential solutions for Indian legal question-answering tasks, highlighting the importance of considering different embedding and QA model combinations for effective performance.

\section{Related Work}
\label{sec:related_work}
Numerous studies, datasets, workshops, competitions, and surveys have been conducted on Chinese \cite{10.5555/3491440.3492202} and Japanese datasets\footnote{\href{https://sites.ualberta.ca/~rabelo/COLIEE2023/}{sites.ualberta.ca/~rabelo/COLIEE2023}}, but there is a lack of research on Indian Legal Question Answering (LQA). To the best of our knowledge, no Indian legal dataset or work specifically addressing the question-answering task has been found. A comprehensive survey by \cite{martinez2023survey} provides an overview of various approaches and techniques used in LQA. The survey discusses different types of legal questions, datasets, and evaluation methodologies employed in the field. Traditionally, most QA systems adopt an extractive-based approach, where relevant context is provided, questions are posed, and the model extracts answers from the given context. In contrast, we tackle the problem using the LLM-based Generative Pretrained Transformer (GPT-3 Davinci model) \cite{brown2020language}, which utilizes the full corpus to generate answers. The Competition on Legal Information Extraction/Entailment (COLIEE) workshop is held annually and focuses on the Statute Law Retrieval Task and Legal Textual Entailment Task \cite{kim2023coliee, rabelo2022overview, rabelo2021coliee, rabelo2020summary, yoshioka2018overview, kim2015coliee}. The retrieval task involves reading a legal bar exam question and extracting a subset of Japanese Civil Code Articles from the entire Civil Code that are relevant to answering the question. The legal textual entailment task involves identifying an entailment relationship, where relevant articles are retrieved given a question. \cite{caballero2022study} focuses on the performance of various transformer-based language models in specialized datasets for the question-answering task in the software development legal domain. \cite{ravichander2019question} address Question Answering for privacy policies by combining computational and legal perspectives. \cite{askari2022expert} proposes methods for generating query-dependent textual profiles for lawyers, which encompass aspects such as sentiment, comments, and recency. They combine these query-dependent profiles with existing expert-finding methods.

\section{Dataset}
\subsection{Documents Collection and Preprocessing}
The dataset comprises thousands of documents pertaining to criminal law, encompassing acts listed at Table \ref{appendix:list_of_acts}. These acts have been obtained from the IndiaCode\footnote{\href{https://www.indiacode.nic.in/}{indiacode.nic.in}} website. Additionally, various articles and blogs related to criminal law have been scrapped from websites such as 
Mondaq\footnote{\href{https://www.mondaq.com/5/India/Criminal-Law}{.mondaq.com/5/India/Criminal-Law}} and LawyersClubIndia\footnote{\href{https://www.lawyersclubindia.com/articles/default.asp?cat_id=7&member_id=&popular=&offset=1}{lawyersclubindia.com/articles}}. The dataset also includes Supreme Court Judgments, scrapped from IndianKanoon\footnote{\href{https://indiankanoon.org/}{indiankanoon.org}}, related to Criminal cases spanning from 1947 to 2020, amounting to a total of 7,221 documents. 

After collecting the documents, a preprocessing step is performed to clean the data. This involves removing line breaks and unnecessary spaces from the text. The processed documents are then saved as text files in a designated directory. Table \ref{tab:dataset_stats} shows the corpus statistics for different criminal article distributions. 

%%%%%%%%%%%%%%%%%%%%%%%%%%%%%%%%%%%%%%%%%%%%%%%%%%
\begin{table}[htbp]
\caption{List of Acts}
\centering
\begin{tabular}{|c|l|}
\hline
\textbf{S.No.} & \textbf{Act} \\
\hline
1 & Indian Penal Code \\
2 & Protection of Children from Sexual Offences Act \\
3 & Criminal Procedural Code \\
4 & Indian Evidence Act \\
5 & Arms Act \\
6 & Information Technology Act \\
7 & Narcotic Drugs and Psychotropic Substances Act \\
8 & Contempt of Courts Act \\
9 & Unlawful Activities Prevention Act \\
10 & Prevention of Money Laundering Act \\
11 & Criminal Procedure Identification Act \\
12 & Extradition Act of 1962 \\
13 & Prisons Act of 1894 \\
14 & Prevention of Corruption Act of 1988 \\
15 & Gram Nyayalayas Act of 2008 \\
\hline
\end{tabular}
\label{appendix:list_of_acts}
\end{table}
%%%%%%%%%%%%%%%%%%%%%%%%%%%%%%%%%%%%%%%%%%%%%%%%%
\begin{table}[h]
\caption{Corpus Statistics for Different Criminal Law Documents Distribution}
\centering
{
\begin{tabular}{|c|c|c|}
\hline
% \textbf{Data} & \textbf{\thead{Average Word\\Count}} & \textbf{\thead{Number of\\Documents}} \\
\textbf{Data} & \textbf{Word Count(Avg)} & \textbf{No. Of Docs} \\
\hline
% \makecell{Criminal\\Judgements} & 4021 & 6942 \\ 
Judgements & 4021 & 6942 \\
\hline
Acts & 28705 & 15 \\
\hline
Articles & 1557 & 264 \\
\hline
\end{tabular}}
\label{tab:dataset_stats}
\end{table}
%%%%%%%%%%%%%%%%%%%%%%%%%%%%%%%%%%%%%%%%%%%%%%%%%%%

\subsection{Test Data}
To assess the performance of different combinations of answer generation and document retrieval models in our legal query answering system, we created a test dataset. This dataset consisted of 50 questions and their corresponding answers, covering various topics in criminal law, such as anticipatory bail, cybercrime, juvenile issues, and sex crimes. These questions were scrapped from the VidhiKarya.Com\footnote{\href{https://vidhikarya.com/free-legal-advice}{vidhikarya.com/free-legal-advice}} website, a trusted online forum for legal discussions.

The answers provided on this website were given by authorized lawyers, ensuring their reliability and accuracy. We used these lawyer-provided answers as the ground truth against which we compared the answers generated by our model.

% \subsection{Prompt for Retrieve Answer}
% Your task is to answer a question as a legal assistant to the best of your abilities, using the context given in the document. If the country is not mentioned in the question, your response should be related to India. You have knowledge of all laws and legal judgments of India. Be detailed in your answer, provide relevant sections and case laws in your answer only if you are confident that they are correct.
% Note that if you do not know the answer, it is acceptable to say "Sorry, I don't know."

% \subsection{Rating Description}
% \begin{enumerate}
%     \item The answer is entirely incorrect or fails to provide any answer.
%     \item The model misunderstood the question and did not offer a relevant response.
%     \item The answer is partly accurate but overlooks essential details.
%     \item A comparable, relevant answer to the ground truth.
%     \item The answer is entirely accurate and relevant, providing a superior response to the expert answer.
% \end{enumerate}

\section{Experimental Setup}
\label{sec:experimental_setup}
\begin{figure}[t]
    \centering
    \includegraphics[width=\linewidth]{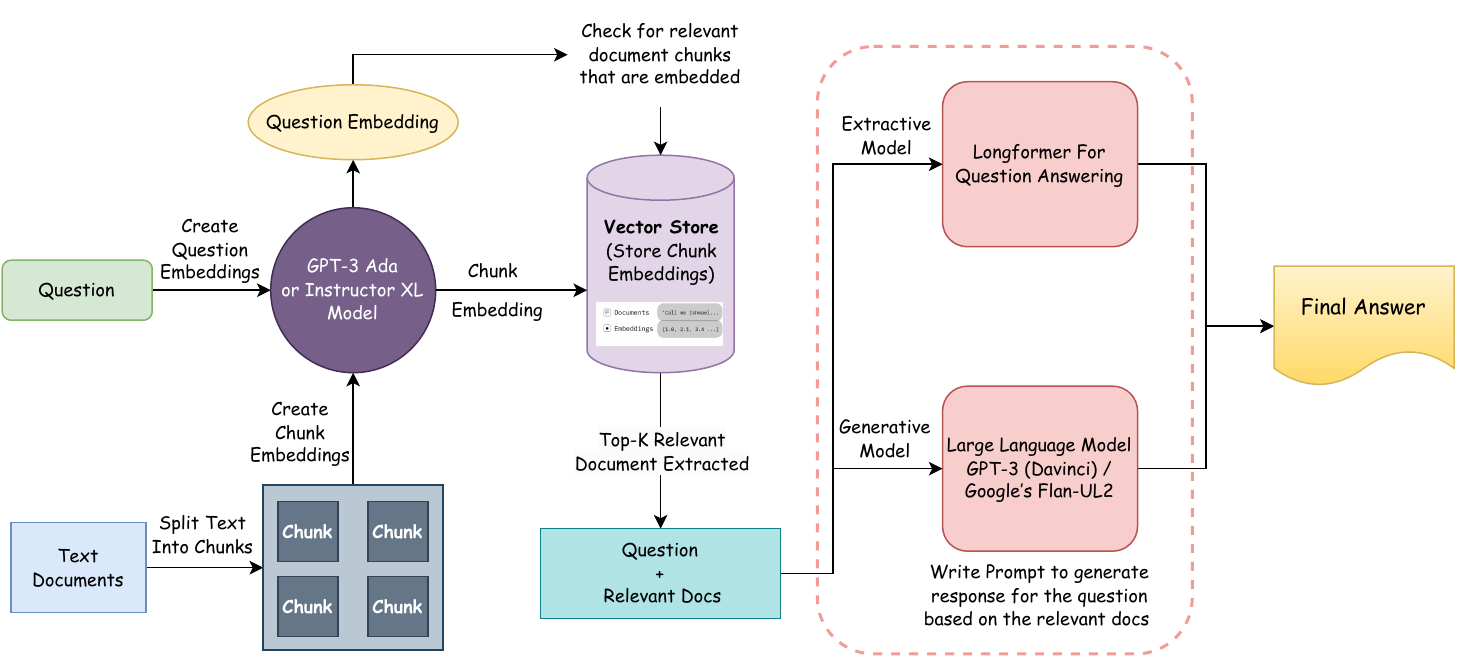}
    \caption{Flowchart of the Legal QA System, integrating Longformer, GPT-3's Davinci, and Flan-UL2 for Extractive and Generative Answer, alongside the inclusion of Pertinent Prompts for Generative models.}
    \label{fig:Flowchart}
\end{figure}

In this section, we provide an overview of the context-based QA system we have developed for the legal field. The system utilizes relevant legal documents in conjunction with user questions, passing them through a Large Language Model (LLM) to generate accurate answers. The experimental setup for our question-answering system can be visualized through the stages depicted in Figure \ref{fig:Flowchart}. 

To retrieve the necessary documents, two methods are compared. The first method utilizes semantic-based retrieval, employing embedding similarity scores to extract the top-k relevant documents. The second method employs the well-established BM25 retrieval algorithm for extracting the top k documents. By comparing the performance of these two retrieval systems, we assess their accuracy in retrieving documents aligned with the corresponding questions.

To facilitate the implementation of these systems, we make use of the LangChain Framework\footnote{\href{https://python.langchain.com/en/latest/index.html}{python.langchain.com}, \href{https://github.com/hwchase17/langchain}{github.com/hwchase17/langchain}}. LangChain is a powerful tool designed specifically for working with LLMs. It addresses the challenge of LLMs lacking domain-specific expertise by preprocessing the text corpus into chunks or summaries. These chunks are then embedded in a vector space, enabling real-time searching for similar chunks when questions are posed. The LangChain Framework simplifies the process of composing these components, allowing for seamless interaction with LLMs. It serves as a valuable tool in scenarios requiring prompt plumbing, such as code and semantic search.

For further insights into the versatile applications of LangChain, we recommend referring to its Documentation\footnote{\href{https://python.langchain.com/en/latest/index.html}{python.langchain.com}} and GitHub\footnote{\href{https://github.com/hwchase17/langchain}{github.com/hwchase17/langchain}} repository.

\subsection{Embedding Based Retrieval System}
Previous sections have covered the initial steps of data collection, preprocessing, and text data dumping, as outlined in the Dataset section. In this section, we will delve into the subsequent steps of our experimental process.
%%%%%%%%%%%%%%%%%%%%%%%%%%%%%%%%%%%%%%%%%%%%%%%%%%%%%%%%%%%%
%%%%%%%%%%%%%%%%%%%%%%%%%%%%%%%%%%%%%%%%%%%%%%%%%%%%%%%%%%%%
\subsubsection{Chunking}
% The process of chunking documents for efficient retrieval involves several steps. Initially, the stored documents are retrieved, and then chunks are created using the CharacterTextSplitter\footnote{\href{https://python.langchain.com/en/latest/modules/indexes/text_splitters/examples/character_text_splitter.html}{python.langchain.com/text\_splitters}} function provided by the Langchain Framework. This function has parameters such as separator, chunk size, and chunk overlap.

% By default, the function uses the separator \verb|'\n\n'| to split the text, but in our case, we have opted to use the separator \verb|'.'|. Additionally, we have set the chunk size to 1000 characters. During the initial splitting stage, the function divides the text based on the specified separator. If a chunk exceeds the specified chunk size, it remains a standalone chunk without further division. However, if a chunk falls below the specified size, it attempts to merge with another chunk to meet the desired chunk size.

% To ensure a seamless flow of information, we have introduced an overlap between the chunks. The overlap parameter, set to 250, means that 250 characters will be shared between adjacent chunks.

% The purpose of chunking is to enhance the retrieval process by focusing on the most relevant segment of a document instead of retrieving the entire document as context for answering a question. The inclusion of overlap ensures that there is no loss of contextual information between adjacent chunks.

The document chunking process for efficient retrieval involves retrieving the stored documents and creating chunks using the Langchain Framework's CharacterTextSplitter\footnote{\href{https://python.langchain.com/en/latest/modules/indexes/text_splitters/examples/character_text_splitter.html}{python.langchain.com/text\_splitters}} function. Chunks are created by splitting the text based on a specified separator, such as \verb|`.'|, with a chunk size of 1000 characters. If a chunk exceeds the chunk size, it remains as is, while smaller chunks may be merged with adjacent ones. An overlap of 250 characters ensures a seamless flow of information between adjacent chunks. Chunking enhances retrieval by focusing on relevant document segments, rather than retrieving the entire document, while maintaining contextual coherence.

\subsubsection{Vector Store Database}
The Vector Store Database is an advanced method for document retrieval that utilizes embedding vectors. It represents documents as vectors in a multi-dimensional space, capturing their semantic meaning and relationships. Through a retrieval algorithm, it compares the similarity between query vectors and document vectors, enabling fast and accurate identification of relevant documents. 

There are various Vector Store Database systems available, including ChromaDB, Pinecone, and DeepLake. For our research, we utilized ChromaDB\footnote{\href{https://docs.trychroma.com}{https://docs.trychroma.com}}, an open-source embedding database. ChromaDB supports local storage and provides convenient wrappers for popular embedding models such as Sentence Transformers, OpenAI's embedding API, and instructor-embeddings. Moreover, it allows the use of custom embedding models. In our system, we employed two different embedding models: OpenAI embedding and Instructor Embedding to create two different embedding databases and use one at a time at the time of retrieval. To retrieve documents, ChromaDB employs a nearest neighbor semantic searching mechanism, ensuring precise retrieval of the desired documents.

\subsubsection{Embedding Generation Model}
As discussed in the previous section, in our document retrieval system, we employ two different embedding models: OpenAI\footnote{\href{https://platform.openai.com/docs/guides/embeddings}{platform.openai.com/docs/guides/embeddings}} and Instructor\footnote{\href{https://huggingface.co/hkunlp/instructor-xl}{huggingface.co/hkunlp/instructor-xl}} embedding. We aim to determine which model helps to retrieve more semantically accurate documents for answer generation.

\underline{OpenAI's `Ada' Embedding:}
OpenAI embeddings utilize a remote server accessed through an API key. While it is not a free model, its charges vary depending on the specific embedding model used, typically billed per thousand tokens. We choose the `ada' model for our system, which costs approximately \$0.0004 per 1000 tokens. Since our dataset contains around 61.6 million tokens, the total cost for creating the embeddings of our database documents amounts to approximately \$24.7. The embeddings generated by the `ada' model have a dimension of 1536.

\underline{Instructor-XL Embedding:}
Instructor-XL is an open-source embedding model available to download from hugging face that has undergone instruction fine-tuning to generate embeddings tailored to any task(e.g., classification, retrieval, clustering, etc.). Notably, Instructor-XL has achieved state-of-the-art performance across 70 diverse embedding tasks. The resulting embeddings from this model have a dimension of 768. To obtain embeddings for our document retrieval system, we instruct the Instructor-XL model to \textit{``Generate embeddings for the document retrieval system."}

\underline{Query Processing and Document Retrieval:}
The legal queries are initially entered into the system during the query processing and document retrieval stage. We utilize the selected large language model (LLM) to obtain the query's embedding, which can be either OpenAI's `Ada' or Instructor-XL. It is crucial to use the same LLM model employed for creating the vector store database. Subsequently, we perform a similarity search using the query embedding, retrieving the top k document chunks (in our system, k is set to 4) from the database. These chunks represent the semantically most similar documents to the input query. Once the document chunks are retrieved, they are passed, along with the query, to the answering system. The answering system utilizes the context provided by the retrieved documents to generate the answer for the query.

\subsubsection{Answer Generation}
In the Answer Generation section of our context-based question-answering system, we utilized three distinct models to compare the quality of generated answers. These models include two generative models, namely OpenAI's GPT-3(Davinci)\footnote{\href{https://platform.openai.com/docs/models/gpt-3}{/platform.openai.com/docs/models/gpt-3}} and Google's Flan-UL2~\cite{tay2023ul2}\footnote{\href{https://huggingface.co/google/flan-ul2}{huggingface/google/flan-ul2}}, as well as an extractive question answering model available on Hugging Face, Longformer~\cite{beltagy2020longformer} base 4096 squad v1 \footnote{\href{https://huggingface.co/valhalla/longformer-base-4096-finetuned-squadv1}{huggingface/valhalla/longformer-finetuned-squad}}.

We first discuss how prompts are used in generative models before going into detail about each answer generation model. Prompts are instructions that help guide generative models to create accurate and relevant answers. They act as guidelines or cues for the models to understand what kind of response is expected and provide appropriate answers to the given questions.

\underline{GPT-3's Davinci:}
% \subsubsubsection{GPT-3's Davinci}
The `Davinci' variant of the OpenAI generative model represents a highly robust GPT-3 model acknowledged for its exceptional proficiency across diverse linguistic tasks. Utilizing the `davinci' variant for answer generation incurs a cost of approximately \$0.02 per 1000 tokens. This variant can handle sequences of up to 4097 words, including the prompt, question, and context. Because it can handle a large number of words, it can understand more information from the relevant document we provide and give good answers to questions. The prompt that we entered into the model along with the query and the relevant documents was \textit{``Your task is to answer a question as a legal assistant to the best of your abilities, using the context given in the document. If the country is not mentioned in the question, your response should be related to India. You have knowledge of all laws and legal judgments of India. Be detailed in your answer, provide relevant sections and case laws in your answer only if you are confident that they are correct.
Note that if you do not know the answer, it is acceptable to say Sorry, I don't know.
Context:\{\} 
Question:\{\}."}

\underline{Google's Flan-UL2:} 
Flan-UL2, an open-source encoder-decoder model based on the T5 architecture, was fine-tuned using `Flan' prompt tuning and dataset collection. The Flan-UL2 checkpoint incorporates a receptive field of 2048 tokens which enhances its suitability for our task. Notably, the model showcases state-of-the-art performance across various natural language processing tasks. Furthermore, it demonstrates superior results in in-context learning, surpassing the performance of the 175B GPT-3 model in zero-shot SuperGLUE. We leveraged the capabilities of Flan-UL2 to generate answers to evaluate its effectiveness against GPT-3's `Davinci.' The prompt that we entered into the model along with the query and the relevant documents was \textit{``Answer the following question using the context by reasoning step by step. If you don't know the answer, just say Sorry, I don't know:
Context:\{\} 
Question:\{\}."}

\underline{Longformer-BASE:}
We utilized Longformer, a BERT-like model designed for handling long documents, for extractive question answering. Specifically, the \textbf{``Longformer-Base-4096 fine-tuned on SQuAD v1"} model was chosen, which is pre-trained and capable of processing sequences with up to 4096 tokens. Given the absence of an open-source model fine-tuned for legal question answering in English, this model was the most suitable choice for our task.

\subsection{BM25 Based Retrieval}
\begin{figure}
    \centering
    \includegraphics[width=\textwidth]{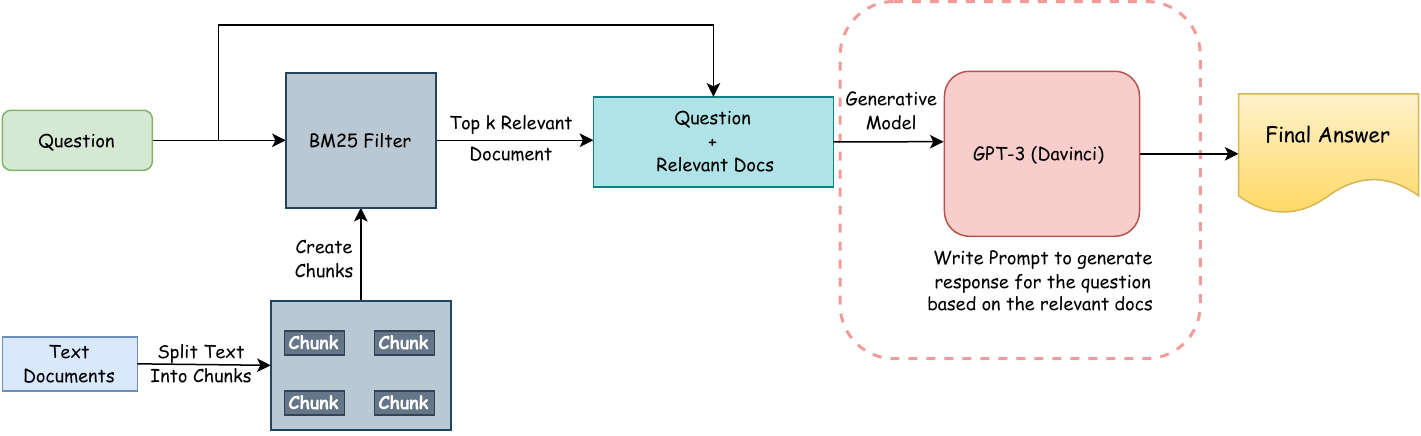}
    \caption{The Diagrammatic Representation of the Question-Answering Framework: Extracting Relevant Documents from the Corpus via BM25 and Generating Answers with OpenAI GPT-3 LLM Model, Augmented with Relevant Prompts.}
    \label{fig:bm25_openai_flowdiagram}
\end{figure}

This section describes the methodology employed for document retrieval and question-answering using the BM25 retrieval algorithm. Figure \ref{fig:bm25_openai_flowdiagram} shows the steps involved in the working of the system. The proposed system aims to retrieve relevant documents from a dataset and generate precise answers to user queries by capitalizing on the strengths of BM25 and advanced language models.

%%%%%%%%%%%%%%%%%%%%%%%%%%%%%%%%%%%%%%%%%%%%%%%%%%%%%%%%%%%%%%%%
%%%%%%%%%%%%%%%%%%%%%%%%%%%%%%%%%%%%%%%%%%%%%%%%%%%%%%%%%%%%%%%%

\subsubsection{Chunking the Document}
The document dataset is divided into manageable chunks, treating each chunk as an independent document. This approach enables efficient processing and retrieval of relevant documents while maintaining the context of the overall dataset.

\subsubsection{Extracting Top-K Relevant Documents}
The user's query or question serves as the input for the retrieval process. Utilizing the BM25 algorithm, the relevance score between the question and each document chunk is computed. This score is determined by considering key factors such as term frequency (TF), inverse document frequency (IDF), and document length normalization. The retrieval system then identifies and extracts the top K(=3) documents with the highest BM25 scores, signifying their relevance to the query.

\subsubsection{Answer Generation System}
The question, along with the top 3 relevant document chunks, is passed as input to the answer generation system. GPT-3's Davinci model, renowned for its language processing capabilities, is employed for generating accurate answers. The relevant documents are utilized as contextual information, providing additional context and supporting evidence for generating comprehensive responses.
By leveraging the BM25 retrieval algorithm and integrating it with advanced language models, the proposed BM25 Retrieval System achieves effective document retrieval and facilitates accurate question-answering.

\section{Experimental Environment}
\label{sec:experimental_environment}
The experimental environment utilized for this research study consisted of the following hardware configurations.

For OpenAI's GPT-3 Davinci, no specific hardware requirements were necessary as it leverages an open API for result generation. Hence, it can be executed on a standard system setup.

Regarding the Flan-UL2 model, loading and running it necessitated a substantial amount of disk space, approximately 40GB, to accommodate the downloaded checkpoint. The system employed for these tasks was equipped with 125GB of RAM, 32 cores CPU, and an NVIDIA A100 40GB GPU, providing ample space and computational resources.

Additionally, the Longformer-Base model was also executed for answer generation on the same system configuration mentioned above.

\section{Performance Metrics}
\label{sec:performance_metrics}
In this section, we outline the various performance metrics utilized to evaluate our legal question-answering system. 

%%%%%%%%%%%%%%%%%%%%%%%%%%%%%%%%%%%%%%%%%%%%%%%%%%%%%%%%%%
% \begin{table*}[t]
% \caption{Average Scores}
%     \centering
%     \resizebox{\textwidth}{!}{%
%     \begin{tabular}{|l|c|c|c|c|c|c|}
%         \hline
%         \multirow{2}{*}{\textbf{Model Name}} & \multicolumn{4}{c|}{\textbf{Context Based Evaluation}}  & \textbf{Semantic Evaluation} & \textbf{Expert Evaluation}\\ \cline{2-7}
%          & \textbf{Rouge-1} & \textbf{Rouge-2} & \textbf{Rouge-L} & \textbf{BLEU} & \textbf{MPNET Score} & \textbf{Rating Score} \\ \hline
%         Ada + Davinci & 0.242 & \underline{0.062} & 0.147 & \underline{0.022} & 0.566 & \underline{3.74} \\ \hline
%         Instructor + Davinci & 0.229 & 0.053 & 0.139 & 0.016 & \underline{0.574} & 3.68 \\ \hline
%         BM25 + Davinci & 0.225 & 0.049 & 0.141 & 0.012 & 0.486 & 2.88 \\ \hline
%         OpenAI (ChatGPT) & \underline{0.267} & 0.052 & \underline{0.158} & 0.010 & 0.561 & 3.54 \\ \hline
%         Instructor + Flan-UL2 & 0.121 & 0.013 & 0.081 & 0.001 & 0.343 & 2.08 \\ \hline
%         Ada + Flan-UL2 & 0.122 & 0.021 & 0.081 & 0.010 & 0.301 & 1.92 \\ \hline
%         Instructor + LongFormer & 0.076 & 0.006 & 0.056 & 0.001 & 0.234 & 1.60 \\ \hline
%         Ada + LongFormer & 0.073 & 0.007 & 0.054 & 0.0 & 0.244 & 1.68 \\ \hline
%     \end{tabular}}
%     \label{tab:average_scores}
% \end{table*}
%%%%%%%%%%%%%%%%%%%%%%%%%%

\subsection{Syntactic Based Method}
The syntactic-based methods include Rouge scores (Rouge-1, Rouge-2, and Rouge-L) and the BLEU Score.

\subsection{Semantic Similarity Based Method}

To assess the semantic similarity between the generated answers and the ground truth, we employed the sentence transformer model, specifically the mpnet ~\cite{jayanthi2021evaluating} base v2 model from HuggingFace\footnote{\href{https://huggingface.co/sentence-transformers/all-mpnet-base-v2}{huggingface/sentence-transformers/all-mpnet-base-v2}}. This model maps sentences and paragraphs to a 768-dimensional dense vector space and can be utilized for tasks such as clustering and semantic search. It is a pretrained MPNET-base model fine-tuned on a dataset containing 1 billion sentence pairs. So this model can give justifiable scores.

% Prior to employing the MPNET model, we attempted to use other available evaluation metrics commonly used for generative models, such as BertScore~\cite{zhang2020bertscore} and KPQA~\cite{lee-etal-2021-kpqa} metric. However, we encountered challenges when utilizing these metrics. BertScore could not provide accurate similarity scores for edge cases, particularly when the model outputted "Sorry, I don't know" as the answer. Despite the model's lack of response, BertScore showed a similarity score of 0.77, prompting us to exclude it from the evaluation process.

% The other metric we tried was KPQA(A Metric for Generative Question Answering Using Keyphrase Weights), designed specifically for evaluating the correctness of the Generative Question Answering system. This metric assigns different weights to each token through keyphrase prediction, aiming to assess whether a generated answer sentence captures the key meaning of the reference answer. The paper demonstrated that the proposed metric exhibits a significantly higher correlation with human judgments compared to existing metrics. We attempted to use the code provided on GitHub\footnote{\url{https://github.com/hwanheelee1993/KPQA}}, which successfully ran on the sample dataset provided. However, when applying it to our outputs, we encountered CUDA-asserted errors that proved challenging to resolve despite our efforts.
Prior to employing MPNET as a substitute model for evaluation, we explored various commonly used evaluation metrics for generative models, including BertScore~\cite{zhang2020bertscore} and KPQA~\cite{lee-etal-2021-kpqa}. However, we faced challenges in utilizing these metrics effectively. BertScore, despite its potential, failed to provide accurate similarity scores for edge cases, particularly when the model outputted responses such as "Sorry, I don't know." Interestingly, even though the model did not produce a meaningful response, BertScore still assigned it a similarity score of 0.77, leading us to exclude it from further evaluation.

We also explored KPQA (A Metric for Generative QA Using Keyphrase Weights) as an evaluation metric. KPQA assigns different weights to tokens based on keyphrase prediction to assess the accuracy of generated answers. Previous studies have shown its higher correlation with human judgments. However, despite successful execution of the code provided on Github\footnote{\url{https://github.com/hwanheelee1993/KPQA}} on their sample dataset, we encountered challenging CUDA-asserted errors when applying it to our own outputs.

\subsection{Human Evaluation}
We incorporated human evaluation, in which law experts assessed the answers generated by our model in comparison to the ground truth. The experts assigned ratings from 1 to 5 based on the following criteria:\\
1. The answer is entirely incorrect or fails to provide any answer.\\
2. The model misunderstood the question and did not offer a relevant response.\\
3. The answer is partly accurate but overlooks essential details.\\
4. A comparable, relevant answer to the ground truth.\\
5. The answer is entirely accurate and relevant, providing a superior response to the expert's answer.

% \begin{enumerate}
%     \item The answer is entirely incorrect or fails to provide any answer.
%     \item The model misunderstood the question and did not offer a relevant response.
%     \item The answer is partly accurate but overlooks essential details.
%     \item A comparable, relevant answer to the ground truth.
%     \item The answer is entirely accurate and relevant, providing a superior response to the expert answer.
% \end{enumerate}

\section{Results and Analysis}
\label{sec:results_and_analysis}
% %%%%%%%%%%%%%%%%%%%%%%%%%%%%%%%%%%%%%%%%%%%%%%%%%%%%%%%%%%
\begin{table*}[t]
\caption{Average Scores}
    \centering
    \resizebox{\textwidth}{!}{%
    \begin{tabular}{|l|c|c|c|c|c|c|}
        \hline
        \multirow{2}{*}{\textbf{Model Name}} & \multicolumn{4}{c|}{\textbf{Context Based Evaluation}}  & \textbf{Semantic Evaluation} & \textbf{Expert Evaluation}\\ \cline{2-7}
         & \textbf{Rouge-1} & \textbf{Rouge-2} & \textbf{Rouge-L} & \textbf{BLEU} & \textbf{MPNet Score} & \textbf{Rating Score} \\ \hline
        Ada + Davinci & 0.242 & \underline{0.062} & 0.147 & \underline{0.022} & 0.566 & \underline{3.74} \\ \hline
        Instructor + Davinci & 0.229 & 0.053 & 0.139 & 0.016 & \underline{0.574} & 3.68 \\ \hline
        BM25 + Davinci & 0.225 & 0.049 & 0.141 & 0.012 & 0.486 & 2.88 \\ \hline
        OpenAI (ChatGPT) & \underline{0.267} & 0.052 & \underline{0.158} & 0.010 & 0.561 & 3.54 \\ \hline
        Instructor + Flan-UL2 & 0.121 & 0.013 & 0.081 & 0.001 & 0.343 & 2.08 \\ \hline
        Ada + Flan-UL2 & 0.122 & 0.021 & 0.081 & 0.010 & 0.301 & 1.92 \\ \hline
        Instructor + LongFormer & 0.076 & 0.006 & 0.056 & 0.001 & 0.234 & 1.60 \\ \hline
        Ada + LongFormer & 0.073 & 0.007 & 0.054 & 0.0 & 0.244 & 1.68 \\ \hline
    \end{tabular}}
    \label{tab:average_scores}
\end{table*}
% %%%%%%%%%%%%%%%%%%%%%%%%%%%%%%%%%%%%%%%%%%%%%%%%%%%
Table \ref{tab:average_scores} presents findings that offer interesting insights into the performance of different models in context-based evaluation, semantic evaluation, and expert evaluation. Each evaluation metric provides information about specific aspects of the models' effectiveness.
\subsection{Context Based Evaluation}
\label{subsec:Context_based_evaluation}
In terms of context-based evaluation, combining Davinci with either Ada or Instructor or BM25 consistently produces good results. The Rouge-1, Rouge-2, Rouge-L, and BLEU scores show that these collaborative models perform well. Among them, the OpenAI's ChatGPT model stands out for its strong understanding of the given context as Rouge-1 and Rouge-L scores for this model are highest among all.

\subsection{Semantic Evaluation}
\label{subsec:Semantic_evaluation}
\begin{figure}
    \centering
    \includegraphics[width=\textwidth]{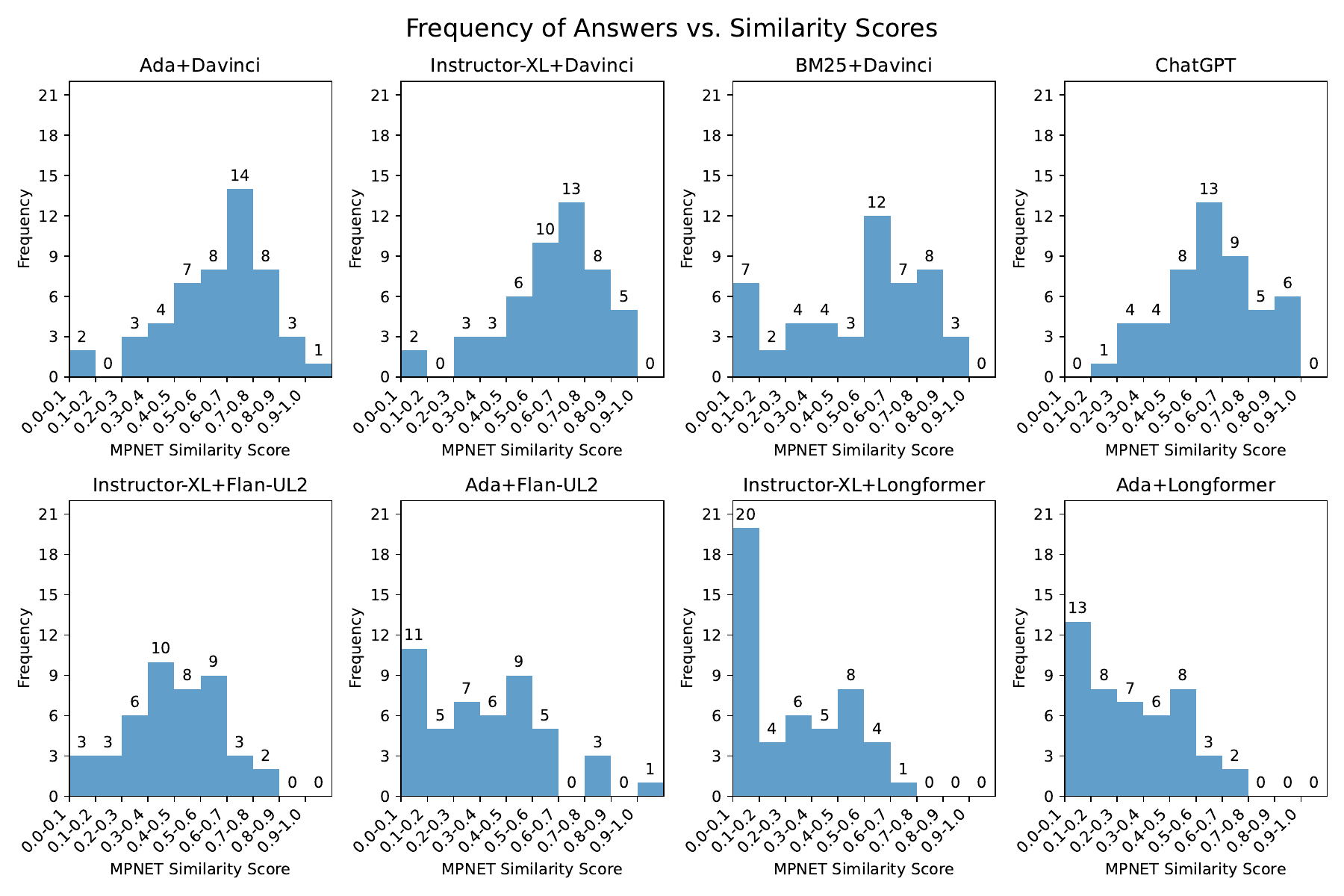}
    \caption{The plot illustrates the relationship between the frequency of answers and their corresponding MPNET similarity scores.}
    \label{fig:frequency_vs_similarity_plot}
\end{figure}

Semantic evaluation reveals how well the models capture the meaning of the prompts. Combining Davinci with Ada or an Instructor shows better performance in terms of MPNET scores, indicating a good grasp of semantic similarity with the original human-generated answer. However, the open-source Flan-UL2, Longformer models, with an instructor or Ada, score lower in semantic evaluation, suggesting that models are not able to produce more accurate answers for the question based on the context.Figure \ref{fig:frequency_vs_similarity_plot} supports our findings, showing that Davinci with Ada or Instructor, and also with BM25, consistently achieve higher MPNET similarity scores (0.6-0.7). Flan-UL2 with Ada or Instructor also performs well, though for a smaller set of questions. In contrast, the Longformer model exhibits the lowest similarity scores, with the majority of question answers falling in the lower range.

Combining Davinci with Ada or an Instructor consistently performs well in contextual understanding, as evidenced by both context-based and semantic evaluations. However, the Flan-UL2 and LongFormer models have room for improvement, particularly in semantic understanding. Notably, the LongFormer model is not suitable for legal question answering at present. Fine-tuning these models on a dedicated legal question answering dataset, though currently unavailable, holds promise for enhancing their performance.

\subsection{Expert Evaluation}
\label{subsec:expert_evaluation}
The evaluation of generative models for the legal question-answering task presents several challenges due to the inherent difficulties in evaluating these models token-wise or syntactically. Moreover, domain-specific fields like medicine and law require a deeper understanding of the subject matter beyond semantic-based methods. Thus, incorporating expert opinion becomes essential to assess the performance of generative models in this context.

% For our evaluation, we utilized a test dataset consisting of 50 public criminal cases and corresponding lawyer answers. The generated answers from the models were presented to a legal expert, who assigned rating scores based on their judgment. The rating scale ranged from one to five, where a score of one indicated that the answer was entirely incorrect or failed to provide any relevant response, while a score of five denoted that the answer was entirely accurate, relevant, and superior to the expert answer.

Upon analyzing the evaluation results presented in Table \ref{tab:expert_scores}, it is evident that the GPT-3 (Ada)-based embeddings performed better than other models in generating answers for the legal QA task. Similarly, the GPT-3 (Davinci) model consistently outperformed other models in terms of answer quality. Notably, employing GPT-3-based models with appropriate prompts for generating embeddings and answers yielded superior results and demonstrated no instances of hallucination.

Furthermore, the evaluation results depicted in Table \ref{tab:expert_scores} highlight that our proposed architecture's answers surpassed those provided by lawyers. This suggests that the generative models, especially those incorporating our proposed architectures, demonstrate a higher level of accuracy and relevance in comparison.

Additionally, we examined the average expert rating scores, as shown in Table \ref{tab:average_scores}, which further support the superiority of generative models over other approaches. Our proposed architectures specifically outperformed ChatGPT-based answers, indicating their ability to generate more reliable responses.
%%%%%%%%%%%%%%%%%%%%%%%%%%%%%%%%%%%%%%%%%%%%%%%%%%
\begin{table}[h]
\caption{Legal Expert Ratings for Various Model Types.}
\centering
\begin{tabular}{|p{3.8cm}|*{5}{c|}}
\hline
\textbf{Model Combination} & \multicolumn{5}{c|}{\textbf{Rating Score}} \\
\cline{2-6}
 & \textbf{1} & \textbf{2} & \textbf{3} & \textbf{4} & \textbf{5} \\
\hline
{Ada + Davinci} & {2} & {7} & {6} & {12} & {21} \\
\hline
Instructor + Davinci& 2 & 7 & 11 & 15 & 15 \\
\hline
BM25 + Davinci & 11 & 11 & 7 & 15 & 6 \\
\hline
OpenAI (ChatGPT)  & 0 & 9 & 13 & 20 & 8 \\
\hline
Instructor + Flan-UL2 & 5 & 36 & 9 & 0 & 0 \\ 
\hline
Ada + Flan-UL2 & 11 & 33 & 5 & 1 & 0 \\ 
\hline
Instructor + LongFormer & 20 & 30 & 0 & 0 & 0 \\
\hline
Ada + LongFormer & 16 & 34 & 0 & 0 & 0 \\
\hline
\end{tabular}
\label{tab:expert_scores}
\end{table}
%%%%%%%%%%%%%%%%%%%%%%%%%%%%%%%%%%%%%%%%%%%%%%%%

% %%%%%%%%%%%%%%%%%%%%%%%%%%%%%%%%%%%%%%%%%%%%%%%%
%%%%%%%%%%%%%%%%%%%%%%%%%%%%%%%%%%%%%%%%%%%%%%%
In summary, the evaluation of generative models for legal question answering reveals valuable insights. Combining Davinci with Ada or an Instructor consistently performs well in contextual understanding, as supported by context-based, semantic, and expert evaluations. 
% This combination showcases superior performance compared to other models, including the OpenAI (ChatGPT) model, and even surpasses human lawyer answers in terms of accuracy, relevance, and reliability. 
However, there is scope for improvement in the Flan-UL2 and LongFormer models, particularly in semantic understanding and alignment with expert judgments. The LongFormer model, in its current state, is not recommended for legal QA as it performs poorly in all the evaluation metrics and is giving incomparable results with the generative models. Fine-tuning these models on a specialized legal QA dataset, although currently unavailable, holds promise for enhancing their performance. These findings highlight the strengths and weaknesses of different models and emphasize the significance of expert evaluation in the assessment of generative models for legal QA tasks.

\section{Conclusion and Future Scope}
\label{sec:conclusion}
In this research article, we have explored the development of an AI for Indian Legal QA (AILQA) system, focusing on the criminal domain. Our study has investigated different embedding and QA model combinations to evaluate their performance in the context of Indian legal question answering. We have leveraged the OpenAI GPT model, specifically the Davinci variant, and incorporated prompts to enable the AILQA system to understand natural language queries and generate accurate answers.

Through empirical analysis and expert evaluations, we have gained valuable insights into the challenges and potential solutions for developing effective AILQA systems. Our findings reveal that the performance of different models varies significantly depending on the embedding and QA model combinations used. The combination of Davinci with Ada or an Instructor consistently demonstrates strong contextual understanding, as evidenced by context-based, semantic, and expert evaluations. In fact, our proposed architectures utilizing GPT-3-based models with appropriate prompts outperform the OpenAI (ChatGPT) model and even surpass human lawyer answers in terms of accuracy, relevance, and reliability.

However, there is room for improvement, particularly in models such as Flan-UL2 and LongFormer, which currently exhibit weaknesses in semantic understanding and alignment with expert judgments. Fine-tuning these models on specialized legal QA datasets, although currently unavailable, holds promise for enhancing their performance. Additionally, incorporating techniques such as Chain-of-Thought prompting (CoT) and few-shot learning can further improve the reasoning ability and interpretability of the AILQA system.

Our evaluation methodology, which includes both syntactic and expert evaluations, provides a comprehensive framework for assessing legal case document QA systems. This approach addresses the open question of how best to evaluate such systems in the legal domain. By involving law practitioners in the evaluation process, we ensure that the AILQA system's performance aligns with the expectations and requirements of legal professionals.

% Overall, this research article contributes valuable insights into the challenges and potential solutions for the development of effective AILQA systems in the Indian legal domain. The findings highlight the importance of considering different embedding and QA model combinations, as well as the significance of expert evaluation, in achieving accurate and reliable legal question answering. These insights pave the way for further research and advancements in the field of legal AI, with the ultimate goal of revolutionizing the way legal professionals interact with case law documents and improving the efficiency and effectiveness of legal research and decision-making.

\bibliographystyle{splncs04}
\bibliography{custom}

\end{document}